\documentclass[runningheads]{llncs}
\usepackage{amssymb}
\usepackage{graphicx}
\newcommand{\C}[1]{\texttt{#1}}

\usepackage{todonotes} 

\newcommand{\R}{\textsuperscript{\textregistered}}

\usepackage{subcaption}
\usepackage{multirow}
\usepackage{url}
\usepackage{cite}
\urldef{\mailsa}\path|joern.ploennigs@ie.ibm.com| 
\urldef{\mailsb}\path|{konstantinos.semertzidis1, fabio.lorenzi1, nandana}@ibm.com| 

\usepackage{listings}

\begin{document}

\mainmatter  

\title{Scaling Knowledge Graphs for Automating AI\newline of Digital Twins}

\titlerunning{Scaling Knowledge Graphs for Automating AI of Digital Twins}

%
%
\author{Joern Ploennigs\inst{1}\orcidID{0000-0002-6320-8891} \and \newline
Konstantinos Semertzidis\inst{1}\orcidID{0000-0002-7040-6706} \and \newline
Fabio Lorenzi\inst{1} \and
Nandana Mihindukulasooriya\inst{1}\orcidID{0000-0003-1707-4842}}
\authorrunning{J. Ploennigs et al.}
%
\institute{IBM Research Europe\\
\email{Joern.Ploennigs@ie.ibm.com},\\\email{\{konstantinos.semertzidis1,fabio.lorenzi1,nandana\}@ibm.com}}
%
%

\toctitle{Scaling Knowledge Graphs for Automating AI of Digital Twins}
\tocauthor{Joern Ploennigs}
\maketitle

\begin{abstract}
Digital Twins are digital representations of systems in the Internet of Things (IoT) that are often based on AI models that are trained on data from those systems. Semantic models are used increasingly to link these datasets from different stages of the IoT systems life-cycle together and to automatically configure the AI modelling pipelines. This combination of semantic models with AI pipelines running on external datasets raises unique challenges particular if rolled out at scale. Within this paper we will discuss the unique requirements of applying semantic graphs to automate Digital Twins in different practical use cases. We will introduce the benchmark dataset DTBM that reflects these characteristics and look into the scaling challenges of different knowledge graph technologies. Based on these insights we will propose a reference architecture that is in-use in multiple products in IBM and derive lessons learned for scaling knowledge graphs for configuring AI models for Digital Twins.
\keywords{Knowledge Graphs, Semantic Models, Scalability, Internet of Things, Machine Learning, Digital Twins}
\end{abstract}

\section{Introduction}
\label{sec:intro}

Semantic models are establishing across industries in the Internet of Things (IoT) to model and manage domain knowledge. They range from driving the next generation of manufacturing in Industry 4.0 \cite{kumar2019ontologies, kalayci2020semantic,bader2020knowledge}, to explainable transport \cite{rojas2021leveraging}, energy savings in buildings for a sustainable future \cite{balaji2018brick, dibowski2020applying}. Their application cumulates in the use of semantic integration of various IoT sensors \cite{rahman2020comprehensive} to automate analytics of the created data \cite{dibowski2020applying,zhou2022machine}.

Digital Twins are one area of applying semantic models. A \textit{Digital Twin} is a digital representation of an IoT system that is able to continuously learn across the systems life cycle and predict the behaviour of the IoT system \cite{ploennigs2018future}. They have multiple uses across the life cycle from providing recommendations in the design of the system, to automating its manufacturing and optimizing its operation by diagnosing anomalies or improving controls with prediction \cite{zheng2021emergence}. The core of a Digital Twin is formed by two tightly interacting concepts. First, an AI model, such as a Machine Learning (ML) or simulation model, that is capable of continuous learning from data and explaining and predicting its behaviour. Second, a \textit{Digital Thread} that is linking these underlying data sources across the systems life cycle \cite{singh2018engineering}. Both approaches interact tightly as the Thread needs to be used to automate the configuration of the AI models to allow to scale their application while results from the AI model should be injected back into the Thread to learn and explain knowledge.

Semantic Knowledge Graph technologies are very well suited for implementing this Digital Thread \cite{akroyd2021universal}. They promise to solve several common challenges from normalizing labelling of the various data sources to being flexible enough to be extended over the life cycle when new applications arise \cite{bone2018toward}. However, scaling knowledge graphs is challenging in its own terms \cite{noy2019industry} and in our practice we experience multiple issues in scaling Digital Threads. Within this paper we will deep dive into this use case and discuss some of the practical issues. We will follow the typical industry workflow for designing and selecting a solution from collecting requirements and defining a test example, to deriving a reference architecture and evaluating final realization options for some large scale examples.
The contributions of the paper are:
\begin{itemize}
    \item \textit{Requirements for Digital Twins}: We collect the requirements for semantic representation of Digital Twins in Section~\ref{sec:requirements}.
    \item \textit{In-Use Experience for Scaling}: We discuss our in-use experience in scaling Digital Twins and propose a reference architecture.
    \item \textit{Benchmark model for Digital Twins}: We define a benchmark model for semantic Digital Twins for an manufacturing example that tests some of the identified requirements in Section~\ref{sec:model}.
    \item \textit{Comparison of KG Technologies}: We compare different knowledge graph technologies for managing the semantic models for Digital Twins in Section~\ref{sec:experiments} including our own semantic property graph.
\end{itemize}

\section{State of the art}
\label{sec:soa}

\paragraph{Knowledge Graphs for Digital Twins:}

There are several examples of applying semantic models for representing Digital Twins \cite{akroyd2021universal,kharlamov2018towards,li2021framework}. Kharlamov et al.\ \cite{kharlamov2018towards} argues for the benefits of using semantic models for digital twins e.\,g.\ to simplify analytics in Industry 4.0 settings. Similarly, Kalayci et al.\ \cite{kalayci2020semantic} shows how to manage industry data with semantic models. Lietaert et al.\ \cite{Lietaert21} presents a Digital Twin architecture for Industry 4.0. Chevallier et al.\ \cite{chevallier2020reference} proposes one for Smart Buildings. Akroyd et al.~\cite{akroyd2021universal} reviews multiple approaches for geospatial knowledge graph for a Digital Twin of the UK. Their work demonstrates the challenges in incorporating data from different domains into one knowledge graph like the heterogenity of data sources. These example represent the use of semantic models for building Digital Twins in different industries that we also see in practise.

\paragraph{Semantic Data Management:} A common goal of using knowledge graphs for Digital Twins is to integrate data from various systems. Established solutions exist for doing this with semantic knowledge graphs that also may integrate external data. Pan et al.\  \cite{pan2018survey} presents a survey of semantic data management systems and benchmarks. The authors classify the systems using a taxonomy that includes native RDF stores, RDBMS-based and NoSql-based data management systems.
Besta et al.\ \cite{besta2019demystifying} provide a classification based on the database technology. Their analysis shows that the different design have various pros and cons. Some of the widely-used generic triple-stores such as OpenLink Virtuoso~\cite{erling2012virtuoso}, Apache Jena, Blazegraph, GraphDB excel on managing RDF data, but, do not scale well in integrating non RDF data. General purpose property graphs like Neo4J or JanusGraph lack intrinsic understanding of semantic models. Multi-modal databases like ArgangoDB or Redis combine a no-sql database with a graph database that allows to manage documents alongside the graph. But, they also suffer from a good understanding of semantic \cite{samuelsen2020approach}. Entris~\cite{endris2019ontario} and Schmidt~\cite{schmid2019using} extend this idea and use semantic models to manage additional data in a data lake. In Section~\ref{sec:requirements} we will discuss  some unique requirements that create challenges in scaling such knowledge graphs. We derive a reference architecture that separation the semantic graph layer from the data layer to scale better to large volumes of data and have federated access to address the Semantic Digital Threads requirements. As shown by our experiments, such design seems to provide better scalability for our use case compared to the other semantic data management approaches.

\paragraph{Benchmarks for Semantic Data:} To validate that the requirements in modelling Digital Twins are unique and evaluate different knowledge graph technologies, we created a new Digital Twin Benchmark Model (DTBM). We compare it against some established benchmarks. The \textit{Berlin SPARQL Benchmark (BSBM)}~\cite{bizer2009berlin} and \textit{Lehigh University Benchmark (LUBM)}~\cite{guo2005lubm} are generic RDF Benchmarks that run a variant of queries on generated datasets. \textit{SP\textsuperscript{2}Bench}~\cite{schmidt2009sp} is based on DBLP library dataset and reflects the social network characteristics of semantic web data. \textit{DBpedia SPARQL benchmark}~\cite{morsey2011dbpedia} uses real queries that were performed by humans and applications against on DBpedia. Additional work reflects the requirements and characteristics of certain domains. \textit{PODiGG}\cite{taelman2019generating} and \textit{GTFS-Madrid-Bench}\cite{chaves2020gtfs} are examples of benchmarks for public transport domain focused on use cases and requirements on route planning on gespatial and temporal transport data. \textit{LSLOD}~\cite{hasnain2017biofed} contains datasets from the life sciences domain and the Linked Data cloud and 10 simple and 10 complex queries that need query federation. \textit{Fedbench suite}~\cite{schmidt2011fedbench} evaluates efficiency and effectiveness of federated SPARQL queries over three different datasets: cross-domain, life science, and SPBenc. We will use BSBM and LUBM in the evaluation in Section~\ref{sec:experiments} as they are very well established and tested for many knowledge graphs technologies and address themselves different RDF characteristics. In addition, we will propose a new benchmark focused on our use case.
\section{Requirements for Semantic Digital Threads}
\label{sec:requirements}

A Digital Thread is linking data from different life cycle stages of a Digital Twin. This starts from design documents such as textual requirements, test specifications, to CAD-files and handbooks that may exist in different formats. During production additional properties may be attached to a Digital Twin such as sensor data from machines, materials used, material providers, and test results. During operation the data collected from the final system is also added. It is often related to asset management data such as fault reports, maintenance work-orders, and replacement histories as well as timeseries data collected from IoT sensors embedded in the systems such as temperature measurements, operational states or alarms. The different datasets that are collected across the life cycle are linked together in the Digital Thread and often analyzed by Machine Learning algorithms to discover and explain anomalies, predict the behaviour of the system and advise people in improved manufacturing and operation of the system.

From this description we can synthesize some \textit{characteristics} to a semantic knowledge graph that can be used to implement such a Digital Thread. 
\begin{itemize}
    \item \textit{C1 - Heterogenous Semantic Types:} The connected data is very heterogenous, representing domainspecific semantic types. A domain ontology can contain thousands of types. For example, the BRICK ontology \cite{balaji2018brick} contains ca.~3.000 classes for modelling smart buildings datasets.
    \item \textit{C2 - Multi-modal Representation:} The data is multi-modal and represented in different formats from timeseries, to binary files, and text documents.
    \item \textit{C3 - Federated Data:} The data is stored and managed in various systems such as complex Continuous Engineering Systems, Asset Management Systems, or IoT platforms.
    \item \textit{C4 - Flexible Hierarchies:} Data is often structured in hierarchical models such as location hierarchies (\C{Country}
    $>$ \C{City} $>$ \C{Factory} $>$ \C{Production\_Line}) and asset hierarchies (\C{Robot} $>$ \C{Arm} $>$ \C{Joint}) that are of flexible depth.
    \item \textit{C5 - Large size:} We see graph sizes often in the range of 100.000 datasets for a mid-size Digital Twin.
    \item \textit{C6 - Composability:} Digital Twins often contain other Digital Twins. For example, a factory twin may contain a robot twin.
    \item \textit{C7 - Lack of semantic knowledge:} We often experience that domain experts do not have deep semantic knowledge. Though, they often understand software engineering concepts like classes and inheritance.
    \item \textit{C8 - Dynamic:} Digital Twins change over their lifetime and so does the Digital Thread. In consequence, the knowledge graph does change regularly bringing in the need to represent time, states and versioning.
\end{itemize}
The \textit{goals} for building the Digital Thread are:
\begin{itemize}
    \item \textit{G1 - Data Linking:} The first goal of the Digital Thread is to link data from various life cycle stages and backend systems together to create an integrated view of the data.
    \item \textit{G2 - Data Contextualization:} The second goal of building Digital Threads is to contextualize the data and understand spatial and functional context to summarize and explain the data.
    \item \textit{G3 - Data Model Normalization:} The third goal is to reduce the heterogeneity of the underlying data and normalize it on: common semantics (C1), a common data modality (C2), and common hierarchical model (C4).
    \item \textit{G4 - Data Access Abstraction:} The next goal is to abstract the access to the underlying data in the federated systems (C3) to allow users to query data by its semantics rather then storage specific IDs like asset or sensor IDs.
    \item \textit{G5 - AI Automation:} The final goal is to automate lifecycle processes like analytics. This is needed as manual configuration of these analytic processes is not possible due to the data size (C5) and regular changes (C8).
\end{itemize}
From these characteristics and goals we can derive a set of requirements:
\begin{itemize}
    \item \textit{R1 - Domain Taxonomies:} From G3 and C1 we can derive the requirement to model both normalized upper ontologies (e.\,g.~\C{Sensor})
    with generic types on the top  and more specific types on lower level of the taxonomy (e.\,g.~\C{Temperature\_Sensor} $\sqsubseteq$ \C{Sensor}).
    \item \textit{R2 - Subsumption:} From C1 and R1 we can derive the requirement to use subsumption in the taxonomies. Taxonomies may have multiple levels, e.\,g.~most sensor tags in BRICK have about 3-5 levels of parents.
    \item \textit{R3 - Inheritance:} From C1, C6, C7 we can derive the requirement to support inheritance of properties from concepts to instances. In practise, we use this heavily to propagate for example units of measurement or ML model configurations.
    \item \textit{R4 - Semantic Data Access:} The solution needs to provide semantic data access according to G4 to the underlying federated data (C3).
    \item \textit{R5 - Backend agnostic:} The system needs to support various data representation (C2) and federated storage solutions (C3) in a hybrid cloud.
    \item \textit{R6 - Flexible Depth Queries:} The hierarchies from C4 provide some means of structuring and querying data. However, the lack of defined structures with fixed query depth requires the use of transitive property chains in queries.
    \item \textit{R7 - Event-based Reasoning:} Reasoning approaches are a good way of automating processes in the knowledge graph for G5 to replicate knowledge for sub-components (C6). The high dynamic of the graph (C8) asks for ways to automate these reasoning steps on graph events, when for example sub-components are added.
    \item \textit{R8 - Guaranteed Consistency:} C1 and C8 mean also that users regularly change the domain taxonomies and there need to be ways to propagate changes in the subsumption taxonomies or the deletion of concepts that keep the graph consistent and not end up with orphaned elements.
    \item \textit{R9 - Element-level Access Control:} The Digital Twin is integrating data from various systems (C3, C6) and needs to support different use cases and user roles (C8). In consequence, a fine grained access control on graph element level is needed \cite{bader2020towards}.
\end{itemize}

Some of these requirements are of common nature for semantic knowledge graphs like R1, R2, R5 and therefore support the applicability of this technology. We consider the requirements R3, R4, R5, R7, R8, R9 more specific for Digital Twins and do not see them in other applications\cite{noy2019industry,Mihindukulasooriya22}.

\section{Architecture for Semantic Digital Twins}
\label{sec:architecture}

\begin{figure}
    \centering
    \includegraphics[width=0.9\textwidth]{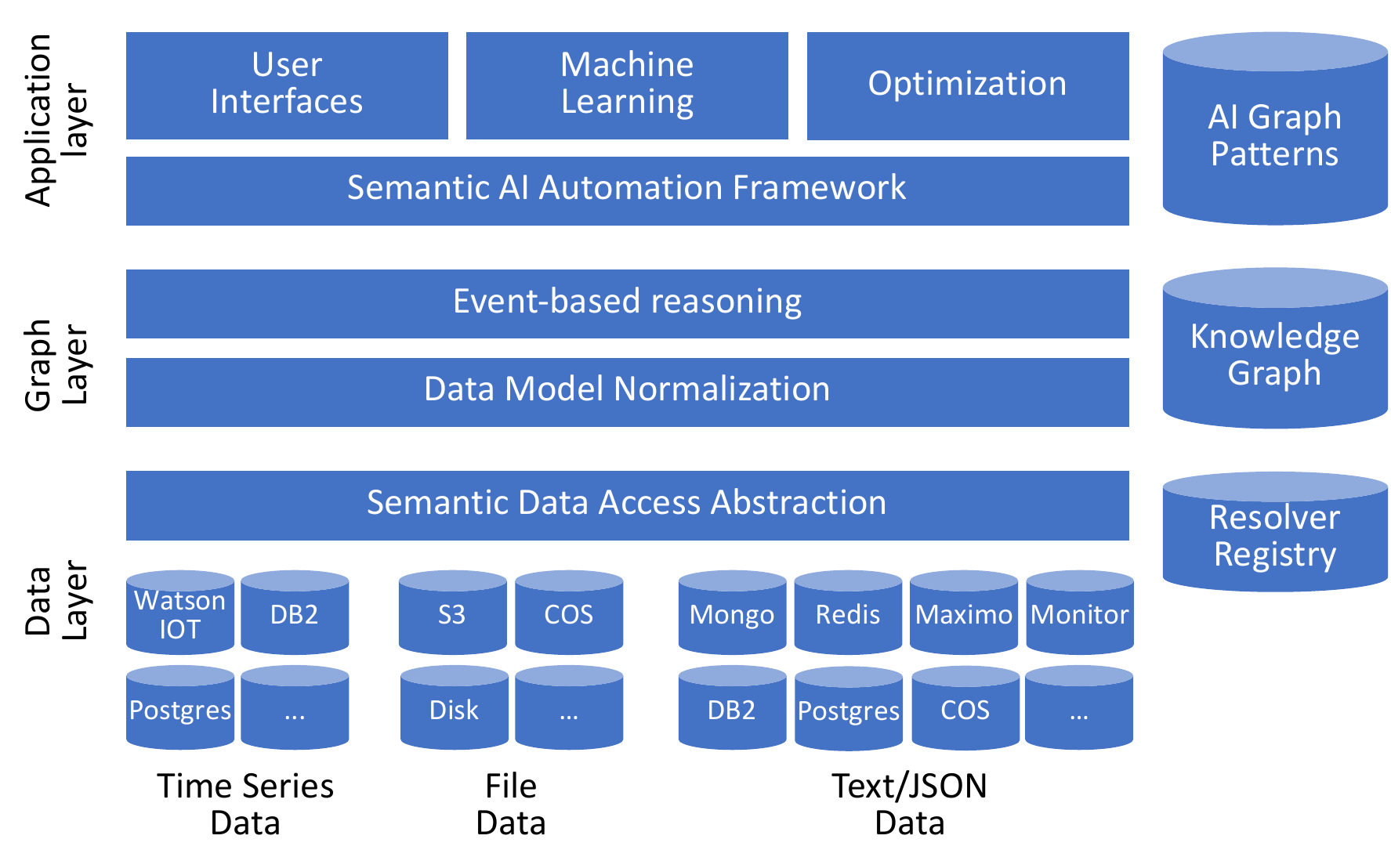}
    \caption{Reference Architecture for Semantic Digital Twin}
    \label{fig:architecture}
\end{figure}

Based on the goals and requirements defined in the last section, we derive a reference architecture for a Semantic Digital Twin in Fig.~\ref{fig:architecture}. We keep the reference architecture on purpose generic as we need to support different backends (C3) and want to give readers implementation options. We realized the reference architecture in our own implementation KITT, which is used in different products like IBM Maximo\R or IBM TRIRIGA\R to integrating data from multiple different solutions in a Digital Threads. The system is deployed and in-use for various customers since multiple years.

The core design of the architecture is a separation of the data layer from the graph layer. The \textit{data layer} links to data from various federated backend systems. They are integrated by a \textit{Semantic Data Access Abstraction Layer}. This is a microservice environment that is implementing for each backend a resolver microservice that maps the backends onto a key/value interface to query the respective data in a normalized way according to G4. We differentiate three types of data representations to address the multi-modality (C2): (i) Timeseries Data (DB2\R, PostgreSQL, Watson IoT\R, $\ldots$); (ii) Binary File Data (Disk, RocksDB, COS\R, $\ldots$); (iii) Textual/JSON Data (MongoDB, Redis, Maximo\R, $\ldots$). A registry allows to dynamically add new resolvers. The mapping to a key/value interface provides a simple interface for implementation of the resolvers and allows us to store the keys as properties in the graph.

The \textit{graph layer} contains the knowledge graph that is linking the meta-data from the various backends (G1). The semantic graph represents the normalized data model (G3) annotated with the associated keys from the data layer. By storing only the data keys in the graph we get a separation of the meta-data graph from the data layer that serves two purposes. First, it keeps the graph manageable in size as the underlying large volume of raw data that often reaches multiple terabytes are kept out of the graph. Second, it helps to solve the federated data problem as clients do not want to move the data out of the master data systems.

We developed our own in-house knowledge graph technology based on our learnings from using over the years different off-the-shelf triple stores and property graphs as backend. Due to the unique requirements, none of them could scale to achieve the desired performance (see Section~\ref{sec:experiments}) and manageability targets. 
Based on our experience with triple stores they present two challenges. First, the graph sizes quickly explode when large numbers of properties (like keys) are added, which are secondary for traversals. Although they provide good support for basic RDF semantics most triple stores do not support subsumption inference out-of-the-box (R2) and rely on rule-based materialization. This not only increases the triple number, but, more importantly it is hard to manage consistency in our production context. When users change the structure of taxonomies it is not easy and efficient to update and delete materialized triples such that a consistent state of the graph can be maintained (R8). Property graphs separate the graph from properties and thus could scale better with large number of properties but they do not necessarily are targeting RDF natively resorting to arbitrary conversion tools to support such scenarios. This lack of semantic understanding inflicts subsumption query performance (Section~\ref{sec:experiments}) and has similar manageability problems of the consistency.

Our knowledge graph is an in-memory semantic property graph called KITT (Keep IoT Trivial). It combines the benefits of semantic understanding of RDF stores and the compact representation of property graphs. It uses an in-memory multigraph representation \cite{ingalalli2018querying,ali2021survey} of nodes and edges decorated with properties. Subsumption relationships are directly represented in-memory and allow for guaranteed consistency (R8) such that subconcepts, instances, or relationships cannot exist without the respective parent concepts or property types. The subsumption relationships can also be directly walked by the reasoner. The reasoner is a traversal-based RDFS+ reasoner that supports transitive queries (R6) alongside event-based reasoning (R7) so that whenever new data instances for a concept are added the reasoner will automatically execute relevant update queries. We use this to automatically instantiate e.\,g.\ analytic functions as discussed in the next section. Another unique aspect of the graph is its support for property inheritance (R3) that is used to propagate properties, like AI function configurations from concepts. It also supports element-level access control (R9) to homogenize access control across the applications. The graph does not provide a SPARQL interface as user often lack the required semantic knowledge (C7) and provides an easier to use structured YAML/JSON based query format.

The \textit{application layer} contains the various solutions that utilize the Digital Twin. They consist of AI tools for machine learning and optimization that are automatically configured from the knowledge graph and of frontend user interfaces. To simplify usage of the platform for domain experts (C7) we provide development frameworks in Java and Python that simplify querying and automation of tasks such as AI pipelines.

Through this architecture we support semantic retrieval of data streams which returns with a single call of the in-built reasoner the data alongside it's semantic context in the knowledge graph. For example, the user can ask for all \C{Power} data attached to a \C{Robot}, with its \C{Workorder} history and the last \C{Image} of a camera monitoring the robot. This query will pull the power consumption data from the IoT platform (Watson IoT\R), the maintenance history from an asset management system (Maximo\R) and images from an object store (COS\R).

\section{Digital Twin Benchmark Model}
\label{sec:model}

The requirement analysis in Sec.~\ref{sec:requirements} identified the main requirements for designing a knowledge graph for Digital Twins. To test and compare different technologies for scaling knowledge graphs we created the Digital Twin Benchmark Model (DTBM) that shares the identified characteristics. Figure~\ref{sec:model} shows the main elements of the model structure.

\begin{figure}
    \centering
    \includegraphics[width=\linewidth]{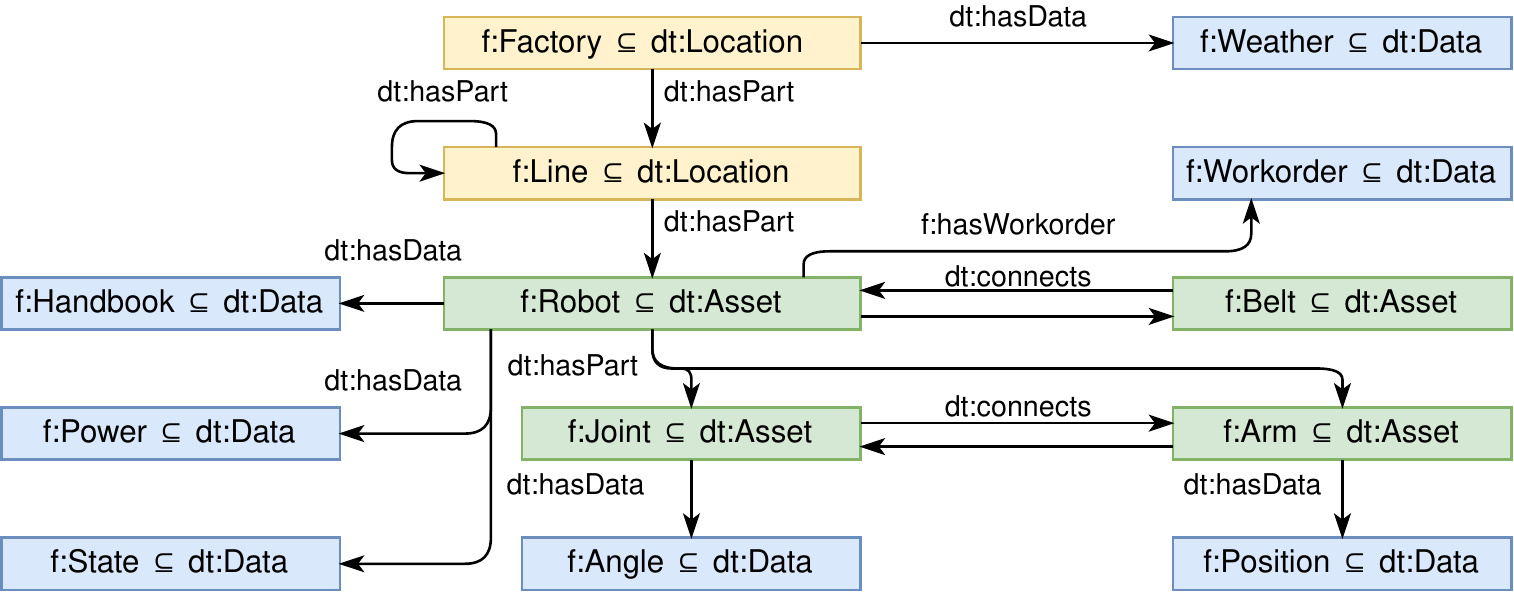}
    \caption{Digital Twin Benchmark Model Structure}
    \label{fig:model}
\end{figure}

The DTBM benchmark is representing an Industry 4.0 example and creates the knowledge graph for the Digital Twin of a production line. The RDF model is split in a core model (\C{dt:*}) and a domain taxonomy (\C{f:*}) according to requirement R1. The core model defines the trinity of \C{Locations}, \C{Asset}, and \C{Data} that is common with other IoT ontologies like BRICK \cite{balaji2018brick}. To keep the benchmark self-contained we refrained from using any upper ontologies. The domain model represents of a location hierarchy in a \C{Factory} $ \sqsubseteq$ \C{Location} with multiple production lines. The production \C{Lines} $ \sqsubseteq$ \C{Location} may have a random depth of 1 to 3 levels to represent flexible hierarchies (C4).

Each production line has multiple \C{Robots} $\sqsubseteq$ \C{Asset} of different types as example for subsumption (R3) with different machine types. Each robot is composed off two to four \C{Joint} $\sqsubseteq$ \C{Asset} and \C{Arm} $\sqsubseteq$ \C{Asset} segments determining the degree of freedom to demonstrate composability (C6).

The robots are connected by \C{Belts} $\sqsubseteq$ \C{Asset} as example for a logistic transport system. It is to note, that the logistic system is introducing a horizontal element cutting across the hierarchical factory structure. These horizontal structures are very common and allow different views on the represented data.

For simplicity we stick to robots and belts in this example, but, they may be any machine and logistic element in a typical production process where a product passes trough a sequence of production steps.

Each robot has different data types attached (C2): a handbook (\C{Data\_File} $\sqsubseteq$ \C{Data}), workorder data (\C{Json} $\sqsubseteq$ \C{Data}), a power meter (\C{Data\_Series\_Numeric} $\sqsubseteq$ \C{Data\_Series} $\sqsubseteq$ \C{Data}), and on/off state (\C{Data\_Series\_Categoric} $\sqsubseteq$ \C{Data\_}-\C{Series} $\sqsubseteq$ \C{Data}) coming from different datastorage systems (C3). Each robot joint has angle measurements and each robot arm has a position measurement to illustrate heterogenity (C1). The belt has power meter and state as well to illustrate that data type may repeat across assets. The factory has weather data as example for a dataset at root level. All elements contain additional data properties to store meta-data such as asset ids or data keys linking to the data in the various backend systems.

The benchmark contains 12 queries. They resemble common type of queries that we see in applications run on the knowledge graph from querying information on specific assets to drive user interfaces and automatically configure and execute ML tasks on the data. The queries are heavily relying on subsumption (R2) and are using primarily generic concepts (\C{Locations}, \C{Asset}, \C{Data}). The queries also use transitive relationships (R6) like the \C{hasPart} relationship that is used to link the factory to lines, down to the robot, and its joints.

\begin{figure}
    \centering
    \begin{subfigure}[t]{0.48\textwidth}
\begin{lstlisting}[language=SPARQL,frame=single,basicstyle=\tiny\ttfamily]
INSERT {
  ?newfunc rdf:type dt:Function .
  ?newfunc dt:hasInputData ?weather .
  ?newfunc dt:hasInputData ?input .
  ?newfunc dt:hasOutputData ?newout .
  ?newout rdf:type f:Data_Power_Pred .
  ?newout dt:hasDataKeySeries "TBD" .
} WHERE {
  ?loc rdf:type dt:Location .
  ?loc dt:hasSeries ?weather .
  ?weather rdf:type f:Data_Weather .
  ?loc dt:hasPart+ ?asset .
  ?asset rdf:type f:Asset .
  ?asset dt:hasSeries ?input .
  ?input rdf:type f:Data_Power .
  BIND(IRI(CONCAT(STR(?asset),"_Pred")) AS ?newout).
  BIND(IRI(CONCAT(STR(?asset),"_Func")) AS ?newfunc).
}
    \end{lstlisting}
    \caption{Query 10 to configure an AI function}
    \label{fig:query10}
    \end{subfigure}
    \hfill
    \begin{subfigure}[t]{0.48\textwidth}
\begin{lstlisting}[language=SPARQL,frame=single,basicstyle=\tiny\ttfamily]
CONSTRUCT {
  ?func rdf:type ?functype .
  ?func dt:hasInputData ?input .
  ?input dt:hasDataKey ?input_key .
  ?func dt:hasOutputData ?output .
  ?output dt:hasDataKey ?output_key .
} WHERE {
  BIND (f:factory1_line_1_robot_1_Func AS ?func) .
  ?func rdf:type dt:Function .
  ?func rdf:type ?functype .
  ?func dt:hasInputData ?input .
  ?input dt:hasDataKey ?input_key .
  ?func dt:hasOutputData ?output .
  ?output dt:hasDataKey ?output_key .
}
    \end{lstlisting}
    \caption{Query 12 to retrieve the sub-graph of an AI function configuration}
    \label{fig:query12}
    \end{subfigure}
    \caption{Query example from the Digital Twin Benchmark Model}
    \label{fig:query}
\end{figure}

Figure~\ref{fig:query} shows two example queries from the benchmark. Query 10 in Figure~\ref{fig:query10} uses SPARQL update to configures a new analytic AI function for each asset (Robot or Belt) that has an associated power timeseries. It uses this power data as input for the analytic function and also adds the weather data as additional input feature from the closes parent owning one (factory). It then assigns an "TBD" data key to the also newly created output timeseries that later will be replaced with a pointer to the newly computed data. The benchmark contains also other examples for configuring AI functions like: Q9 - to aggregate across a location hierarchy or Q11 - to aggregate all instances by asset type. All queries follow the same template of utilizing a graph pattern (WHERE) to identify deployment locations for the AI functions with relevant inputs and a creation (INSERT) part to materialize the respective AI function. Their configurations could be added to the (INSERT) section \cite{ploennigs2014adapting} or are in our case inherited from their type (R3).

Query 12 in Figure~\ref{fig:query12} retrieves the sub-graph for one of these created configurations that contains the inputs and outputs for computing the prediction in an ML job including the data keys. In the given example, this would be a new AI function to compute the prediction of the power meter at robot 1 on line 1 that uses as input the power meter history and the weather data and the newly created output.

\section{Experiments}
\label{sec:experiments}

The goal of our experimental evaluation is threefold. First, we want to illustrate the unique challenges that scaling knowledge graphs for digital twins create across various knowledge graph technologies. Second, we want to prove that the proposed Digital Twin Benchmark Model has different behaviour than other benchmarks to verify the need for another model and explore the different characteristics these benchmarks expect of the knowledge graph. Last, we want to validate the benefits of semantic knowledge graphs at the example of our implementation in comparison to established triple stores and property graphs across different benchmarks. \\

\noindent\textbf{Knowledge Graphs:}
The first three knowledge graph technologies we evaluate are Blazegraph, GraphDB, and Virtuoso all of which are triple stores that expose SPARQL endpoints. The fourth system is Neo4j as most common property graph. The last one is KITT which is our semantic property graph. The selection is based purely on the availability of information on how to run BSBM and LUBM for these backends.

We used the same workflow to set the systems up, load the data into them, and evaluate their performance. Specifically, we used a containerization version for each graph without following any optimization strategy. The purpose is to replicate a cloud environment where a user will not have access to optimize the backend systems nor the knowledge (C7). \\

\noindent\textbf{Benchmarks and Datasets:}
We present the results of two well known RDF benchmarks namely BSBM\cite{bizer2009berlin} and LUBM\cite{guo2005lubm} as well as our proposed Digital Twin Benchmark Model (DTBM). 
Each of them generates RDF datasets which can easily be loaded into the triple stores. For Neo4J we use the neosemantics plugin to import RDF and had to write custom Cypher queries to support all subsumption alternatives. For KITT we use our own RDF loader hat converts RDF in a class and instance model.
More details about the benchmarks:
\begin{enumerate}
    \item The Berlin SPAQRL Benchmark (BSBM)\cite{bizer2009berlin} is built around an e-commerce use case and it is scaled via the number of the products. It creates a set of products offered by different vendors and reviewed by consumers. In general, the generated datasets are property heavy and consist of subsumption hierarchies and their depth depends on the chosen scale factor.
    For evaluating the different systems, BSBM uses a query generator which generates queries that emulate the search and navigation pattern of a consumer looking for a product.
    \item The Lehigh University Benchmark (LUBM)\cite{guo2005lubm} was developed to evaluate the performance of the semantic web knowledge base systems with respect to use in large OWL applications. It generates a dataset for the university domain and its query evaluation uses subsumptions and recursive queries such as transitive closure.
    \item The Digital Twin Benchmark (DTBM) was introduced in the last section. It is subsumption heavy with transitive queries of flexible depth to query information and configure AI functions.
\end{enumerate}

\noindent The characteristics of the datasets generated by the different benchmarks are summarized in Table \ref{table:characteristics}. 
For the BSBM dataset the real "Dataset scale" factor is the value the reported value multiplied by 1,000. For example, S2 in BSBM refers to 2,000 products. BSBM is  evaluated with 16 queries in parallel in Java, while LUBM and DTBM are evaluated with single-threaded sequential queries in Python. We do not run parallel tests to not skew results with Pythons bad parallelization behaviour.

\begin{table*}[ht!]
\centering
\resizebox{1.0\textwidth}{!}{
\begin{tabular}{c|cccc|cccc|cccc}
\hline
\textbf{Dataset}
 & \multicolumn{4}{c}{\textbf{BSBM}}  & \multicolumn{4}{|c|}{\textbf{LUBM}}   & 
 \multicolumn{4}{c}{\textbf{DTBM}}                                                                                        \\ 
 \textbf{scale} & \textbf{triplets} & \textbf{nodes} & \textbf{edges} & \textbf{props} & \textbf{triplets} & \textbf{nodes} & \textbf{edges} & \textbf{props} & \textbf{triplets} & \textbf{nodes} & \textbf{edges} & \textbf{props} \\ 
\hline
S2 & 377,241 & 36,871 & 102,315 & 37,022 & 288,894 & 38,349 & 113,463 & 36,834 & 114,177 & 35,702 & 41,698 & 35,601\\
S5 & 1,711,567 & 152,965 & 454,334 & 153,294 & 781,694 & 102,383 & 309,393 & 100,018 & 283,785 & 88,895 & 103,874 & 88,713\\
S10 & 3,738,188 & 315,148 & 982,502 & 315,477 & 1,591,643 & 207,441 & 630,753 & 203,625 & 570,232 & 178,118 & 208,224 & 177,801\\
S20 & 7,749,994 & 632,447 & 2,018,821 & 632,776 & 3,360,686 & 437,570 & 1,332,029 & 430,559 & 1,136,634 & 356,087 & 416,285 & 355,500\\
S50 & 17,571,059 & 1,593,382 & 5,132,438 & 1,594,113 & 8,317,905 & 1,082,833 & 3,298,813 & 1,067,023 & 2,844,499 & 890,624 & 1,041,300 & 889,227\\
S100 & 35,159,904 & 3,196,023 & 10,104,195 & 3,198,034 & 16,753,468 & 2,179,781 & 6,645,928 & 2,148,846 & 5,693,601 & 1,781,866 & 2,083,488 & 1,779,119\\
\hline
\end{tabular}}
\caption{Dataset Characteristics with their number of elements}
\label{table:characteristics}
\end{table*}

\noindent\textbf{Query Configuration:} In BSBM, we use the default BSBM test driver \cite{bizer2009berlin} which executes sequences of SPARQL queries over the SPARQL protocol against the system under test. 
In order to emulate a realistic workload, the test driver can simulate multiple clients that concurrently execute query mixes against the test system. 
The queries are parameterized with random values from the benchmark dataset, in order to make it more difficult for the test system to apply caching techniques. 
The test driver executes a series of warm-up query mixes before the actual performance is measured in order to benchmark systems under normal working conditions.
To make a fair comparison, we implemented our own test driver that follows a similar design to evaluate Neo4j and KITT. Note, BSBM test driver measures the response time excluding the time needed to deserialize the response and it puts a constraint to the result size. 

In LUBM and DTBM, we use our own query executor in Python to evaluate the performance of the various systems. We execute each query 500 times and we report the average response time including the deserialization of the response. 
We also place a constraint to the result size limiting the result to 1,000 records to put the focus on the query execution time and not the serialization time. 

Overall, all benchmarks use SPARQL queries; corresponding Cypher queries for Neo4j or YAML queries for KITT. We optimize the representation for each graph and e.\,g.\ directly represent properties in Neo4j and optimize queries accordingly to e.\,g.\ activate subsumption support (RDFS+) for all RDF stores. All benchmarks have been executed on a Linux machine with a AMD Ryzen 3959X processor and 64\,GB memory and SSDs with the backends running on Docker.

\begin{figure}[hbt]
     \centering
     \hspace*{-0.5cm}
     \includegraphics[width=1.12\linewidth]{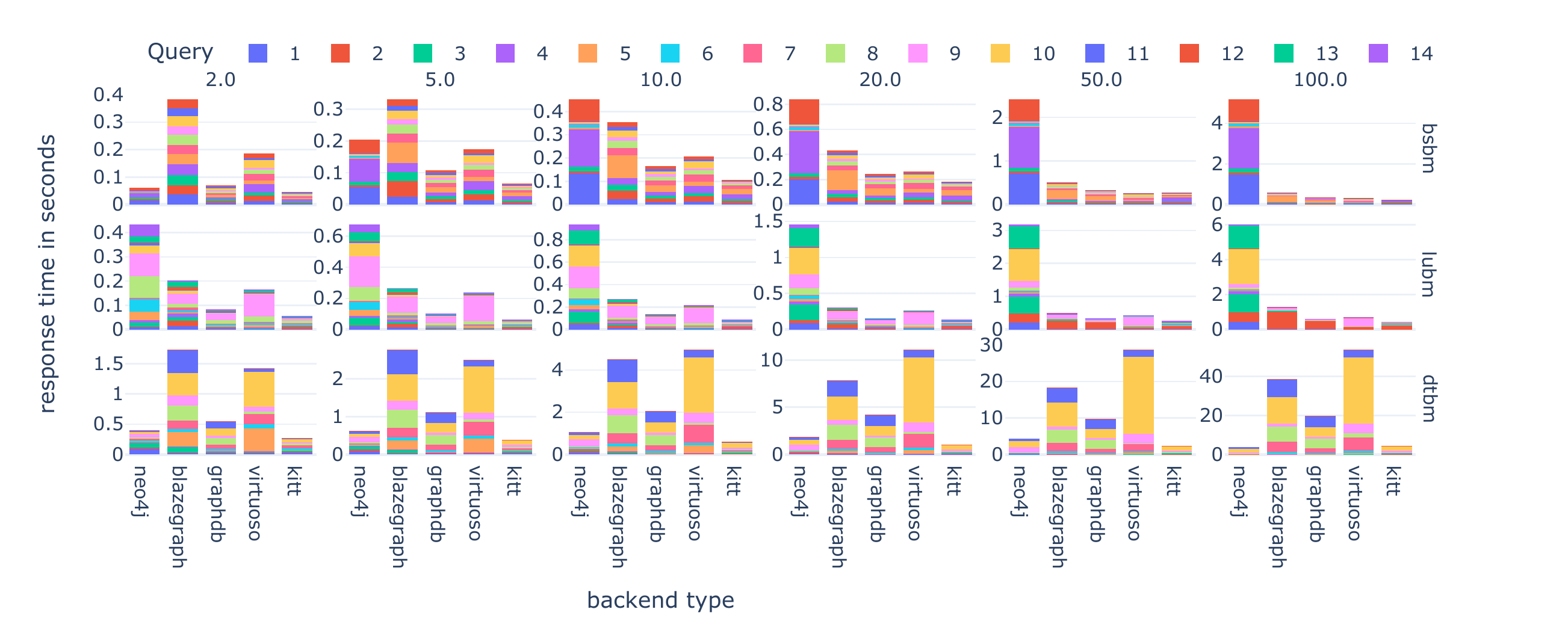}
     \caption{Query response times for different sizes (columns) and benchmarks (rows)}
     \label{fig:query_time}
\end{figure}

\subsection{BSBM evaluation}
Fig.~\ref{fig:query_time} summarizes the average response time for all benchmarks in seconds. We will first discuss the BSBM results in the first row. For this one all queries run in less than a second even for the big graphs.
We observe that Neo4j performs well only for small graphs and worse for larger. It seems that it cannot understand the semantics hidden in the dataset and it does not perform well when queries use subsumption.
Comparing the triple stores, we find that GraphDB has the lowest response times while Blazegraph has the worst performance. It seems that the queries using filtering by multiple properties such as Query 5 are costly for the RDF triple stores.
KITT outperforms the other knowledge graphs in all datasets slightly. This shows its ability as a semantic property graph to understand the model and the semantics contained in the dataset.

\subsection{LUBM evaluation}
The results for LUBM on the second row in Fig.~\ref{fig:query_time} show about the same characteristics as BSBM. However, we observe that Neo4j is not able to perform well for any size. This is due to the fact that LUBM uses many subsumptions and recursive queries and Neo4j seems unable to process them efficiently.
Furthermore, we observe for the different graphs that the ratio some queries have in the stacked bars change with the experiment sizes. This hints that the knowledge graphs struggle with different queries characteristics when a graph grows. For example, Query 9 is the most expensive for triple stores up to dataset size S20, and then we see an increase in response time for Query 2. 
This can be explained by the fact that Query 2 fetches for each department of a university, all the graduate students and searches for those who have obtained their undergraduate degree from the given university. This breadth exploration becomes very costly in larger graphs because the number of students becomes huge and many invalid pathes need to be evaluated before they are dropped from the result set. In addition, Query 9 returns students who have taken a course of their faculty advisor which also leads to multiple traversals of the hierarchy that links advisors to students through a course. Although the query is of a similar structure to Query 2, it is less costly in large graphs due to the smaller search space.
Finally, we observe that KITT again surpasses all knowledge graphs and it can produce the result in a few milliseconds. This again shows the benefits of an in-memory semantic property graph that can directly walk the subsumption taxonomy from high-level concepts to instances and their transitive relationships.

\subsection{DTBM evaluation}
The results for our proposed benchmark illustrate first the bad scaling behaviour of traditional knowledge graphs in a Digital Twin scenario as discussed in the beginning. While common execution times for BSBM and LUBM were usually under one second for the triple stores they grow to multiple seconds for some queries for larger graphs, which is not acceptable for driving user interfaces. This magnitude differences in response times for DTBM is particularly notable as that the size in triples of the dataset size S100 is only 1/3 of LUBM and 1/6 of BSBM (Table~\ref{table:characteristics}).

The second notable aspect is that the performance characteristics between the triple stores change. While Blazegraph was performing worst for BSBM and LUBM it is now Virtuoso that has the largest response times. It is particularly struggling with Query 10 (Figure~\ref{fig:query10}) and actually failed out-the-box for S100 due to too large implication space.

Along the same line it is now Neo4J that overtakes the triple stores in the benchmark. It probably benefits from the lower number of nodes and the property representation. It is notable that across all benchmarks it scales constantly along to the number of nodes. However, we have to note that we had to write custom Cypher queries for Neo4J to support all subsumption alternatives for all use cases. In production this is not feasible because taxonomies are domain specific (R1) and change (C8).

KITT outperforms the other graph stores also for this scenario. This demonstrates the generalizable benefit of a semantic property graph across multiple use case from RDF specific benchmarks like BSBM or LUBM to the digital twin use case addressed in DTBM.

\section{Summary}
In this paper we share our experience in scaling semantic knowledge graphs for automating AI of Digital Twins in the IoT domain. We analyze the unique characteristics, goals and requirements of the use case that uniquely links semantic models to multi-modal data management across different federated data systems.

We derive from this a reference architecture for a Digital Twin knowledge graph solution that we use in multiple products. It separates the knowledge graph from the underlying data in a micro-service environment that is easy to scale and extend.

To enable the community to evaluate different knowledge graph technologies for this architecture we open source a new Digital Twin Benchmark Model (DTBM) that represent the specific requirements of the domain. The DTBM generator creates an Industry 4.0 production line and contains multiple queries that we use in production. We demonstrate in some query examples how AI functions can be automatically configured from the semantic graph.
We execute the benchmark for different knowledge graph technologies and compare it to the well established BSBM and LUBM. The result highlight the different behaviour of DTBM in comparison to BSBM and LUBM and substantiate the need of the new benchmark. They also illustrate the challenges in scaling knowledge graphs for Digital Twins that already show by a magnitude larger response times for even smaller graphs. It may be possible to optimize the parameters, indexes and queries for the different knowledge graph technologies, but, given the domain specificity (R1) and dynamic nature (C8) of these models and queries this is not feasible in production for a general purpose cloud service where users have neither access to these configurations nor the expertise (C7).

That our own knowledge graph KITT shows the best performance across all benchmarks demonstrates the advantage of semantic property graphs that combine benefits of RDF and property graphs and support subsumption and transitivity out of the box. We would like to see more graphs that address this need and not specialize on one or the other. For that purpose we open sourced the script for creating the Digital Twin Benchmark Model as well as the used benchmark configuration such that other can replicate and extend our results.

\paragraph{Supplemental Material Availability:} The source code and queries for DTBM as well as the used configurations for the benchmark are available at\newline \url{https://github.com/IBM/digital-twin-benchmark-model}.

\end{document}